\documentclass[10pt, a4paper, onecolumn]{article} 

\usepackage[english]{babel} 

\usepackage{float}

\usepackage{microtype} 

\usepackage{amsmath,amsfonts,amsthm} 

\usepackage[svgnames]{xcolor} 

\usepackage[hang, small, labelfont=bf, up, textfont=it]{caption} 

\usepackage{booktabs} 

\usepackage{lastpage} 

\usepackage{graphicx} 

\usepackage[boxruled, linesnumbered, portuguese]{algorithm2e}

\usepackage{blindtext}

\usepackage[colorlinks=true,linkcolor=blue,citecolor=red,urlcolor=black,bookmarks=true,pdfstartview=FitB]{hyperref} 

\usepackage{enumitem} 
\setlist{noitemsep} 

\usepackage{sectsty} 
\allsectionsfont{\usefont{OT1}{phv}{b}{n}} 

\usepackage{geometry} 

\usepackage[T1]{fontenc} 
\usepackage[utf8]{inputenc} 

\usepackage{XCharter} 

\usepackage{fancyhdr} 
\fancypagestyle{firstpage}{ 
	\fancyhf{}
}

\newcommand{\authorstyle}[1]{{\large\usefont{OT1}{phv}{b}{n}\color{DarkRed}#1}} 

\newcommand{\institution}[1]{{\footnotesize\usefont{OT1}{phv}{m}{sl}\color{Black}#1}} 

\usepackage{titling} 

\newcommand{\HorRule}{\color{DarkGoldenrod}\rule{\linewidth}{1pt}} 

\pretitle{
	\vspace{-30pt} 
	\HorRule\vspace{10pt} 
	\fontsize{23}{23}\usefont{OT1}{phv}{b}{n}\selectfont 
	\color{DarkRed} 
}

\posttitle{\par\vskip 15pt} 

\preauthor{} 

\postauthor{ 
	\vspace{10pt} 
	\par\HorRule 
	\vspace{20pt} 
}

\usepackage{lettrine} 
\usepackage{fix-cm}	

\newcommand{\initial}[1]{ 
	\lettrine[lines=3,findent=4pt,nindent=5pt]{
		\color{DarkGoldenrod}
		{#1}
	}{}%
}

\usepackage{xstring} 

\newcommand{\lettrineabstract}[1]{
	\StrLeft{#1}{1}[\firstletter] 
	\initial{\firstletter}\textbf{\StrGobbleLeft{#1}{1}} 
}

\title{ALT\thanks{\url{https://legibilidade.com/}}: A software for readability analysis of Portuguese-language texts } 

\author{
	\authorstyle{Gleice Carvalho de Lima Moreno\textsuperscript{1,3}, Marco P. M. de Souza\textsuperscript{2}, Nelson Hein\textsuperscript{3}, Adriana Kroenke Hein\textsuperscript{3}} 
	\newline\newline 
	\textsuperscript{1}\institution{Departamento de Ciências Contábeis, Universidade Federal de Rondônia, 76801-974, Porto Velho, Rondônia, Brasil}\\
	\textsuperscript{2}\institution{Departamento de Física, Universidade Federal de Rondônia, 76900-726, Ji-Paraná, Rondônia, Brasil}\\
	\textsuperscript{3}\institution{Programa de Pós-Graduação em Ciências Contábeis, Universidade Regional de Blumenau, 89030-903, Blumenau, Santa Catarina, Brasil}\\ 
}

\date{\today} 

\begin{document}

\maketitle 

\thispagestyle{firstpage} 


\lettrineabstract{In the initial stage of human life, communication, seen as a process of social interaction, was always the best way to reach consensus between the parties. Understanding and credibility in this process are essential for the mutual agreement to be validated. But, how to do it so that this communication reaches the great mass? This is the main challenge when what is sought is the dissemination of information and its approval. In this context, this study presents the ALT software, developed from original readability metrics adapted to the Portuguese language, available on the web, to reduce communication difficulties. The development of the software was motivated by the  theory of communicative action of Habermas, which uses a multidisciplinary style to measure the credibility of the discourse in the communication channels used to build and maintain a safe and healthy relationship with the public.}
\newline\newline
\noindent ------------------
\newline

\lettrineabstract{No estágio inicial da vida humana a comunicação, vista como um processo de interação social, foi sempre o melhor caminho para o consenso entre as partes. O entendimento e a credibilidade nesse processo são fundamentais para que o acordo mútuo seja validado. Mas, como fazê-lo de forma que essa comunicação alcance a grande massa? Esse é o principal desafio quando o que se busca é a difusão da informação e a sua aprovação. Nesse contexto, este estudo apresenta o software ALT, desenvolvido a partir de métricas de legibilidade originais adaptadas para a Língua Portuguesa, disponível na web, para reduzir as dificuldades na comunicação. O desenvolvimento do software foi motivado pela teoria do agir comunicativo de Habermas, que faz uso de um estilo multidisciplinar para medir a credibilidade do discurso nos canais de comunicação utilizados para construir e manter uma relação segura e saudável com o público.}


\section{Introduction}

Jürgen Habermas is a German philosopher and sociologist from the Frankfurt School who  worked hard to study democracy by devoting extensively to the theory of communicative action published in 1981. This theory emphasized the way in which communication should occur, by multidisciplinary treating the credibility in the relationship between the system (economic and political) and the world of life (represented by the people’s prior knowledge, standards  in society and knowledge from culture). Based on the individual’s innate language, he described that dialog should occur freely, with communicative rationality and making critical analysis in this interaction in order to reach the essential and apex of the communication \cite{Habermas}.

In view of this, influenced by Habermas's theory of communicative action, this work was developed to reduce the flaws that prevent the understanding of communications written in Portuguese. The most common flaws are the absences of objectivity, clarity, and simplicity.

Communication is the main driver who intervenes in the search for a better relationship between people or groups of people. Through dialog (written or oral form), it is important to establish a transmission of information backed by ethics and morals in order to reach persuasion. Following these criteria, the prospect of achieving credibility and, consequently, the concordance of the collectivity is greater.

However, this is not always the case. Written communications are often aimed at a specific audience. Thus, the use of complex words (common to the group) and long sentences is the most common, making reading difficult and preventing the understanding of readers that are part of other groups. Therefore, the message does not reach the expected range, and it does not meet a greater number of people (either layperson or specialist). It is almost a game of bad luck with errors and  trials \cite{Gregoire}.

In this sense, Habermas found, through the theory of communicative action, that the terms used in a communication must follow four pretensions of validity (intelligibility, sincerity, normative correction and truth) in order to reach the summit (credibility) at this important and necessary stage of human interaction.

The pretense of intelligibility corresponds to the communication process that has been clearly carried out, thus allowing for an easy understanding of what has been declared in order to reach consensus between the parties. Thus, this claim refers to the indicator of comprehensibility, which occurs when the degree of effectiveness of communication is achieved. To measure it, there are several textual readability metrics developed over the last decades. We will deal with some of them in this Article.

As for the claim of sincerity, it is the disclosure of information in a detailed manner, presenting honestly what has been done or has not been done. To accurately measure this indicator, keywords should be considered for the purpose of assessing whether the author(s) dealt with or dealt with in detail the subject to which the text is intended.

Regarding the claim of regulatory correction, it has as its principle the compliance with established legal rules or standards, dealing with the adequacy of the reports in relation to a specific situation (environmental, social, cultural, among others) or considering the reader's ability to respond to what is proposed  \cite{Baalouch}. This indicator is measured by means of the readability metrics, which indicate the intended audience for the communication.

The claim of truth, however, deals with the availability of reliable information, always considering the truth of the facts. As stated, reliability is fundamental to the communication process, being obtained from the relationship between the guidelines observed and suggested guidelines based on pre-established standards (standards and rules).

From the Habermas’ theory, it has to be said that, if these claims or claims are present, the communication existing in the relationship between the system (government and market) and the world of life (subjective, normative and objective) will be more solid.

This has been seen as a major challenge for humanity, by influencing the communication process in its various relations. As examples, we can cite the relationship between government and taxpayers; between company and society; between manager and collaborators; between doctor and patients; between scientist and public; and so many other cases. In order to attenuate the imperfections in the communication process, particularly in the written form, some studies have been highlighted because they are pioneers in this field of study.

The field of study referred to relates to readability, which aims to analyze the difficulty understanding a text. Some studies, such as Flesch  \cite{Flesh} in 1948, Gunning \cite{Gunning} in 1952, Smith and Senter  \cite{Smith} in 1967, Coleman and Liau  \cite{Coleman} in 1975, Kincaid  \cite{Flesch-Kincaid} and collaborators in 1975, and Gulpease  \cite{Gulpease} em 1988, in addition to others with scientific evidence, were important for proposing solutions to measure the degree of reading difficulty.

Thus, in this work we present the ALT software: Textual Readability Analysis  \cite{ALT}, a tool developed to measure textual readability indexes of Portuguese texts, using formulas adapted for that language from the originals. Within Habermas's theory of communicative action, readability indices can be used when the objective is to obtain quantitative data of the pretensions of intelligibility and normative correction. In addition, the the claim of sincerity, which seeks to measure the completeness of the text, observing the frequency with which the chosen keywords have been mentioned, is also proposed in the software.

That said, the ALT program was built to meet two needs:

\begin{enumerate}
	\item To enable the analysis of textual readability for texts written in Portuguese.
	
	\item To fill an existing gap in the scientific environment, since researchers from several areas develop studies focused on textual readability in Portuguese and end up making use of international software based on readability indexes not suitable for this language.
	
\end{enumerate}

Still within the second point of the list above, it should be noted that even recent works, with four years or less of publication, used textual readability indices in their original languages (English). Among these studies, it is possible to mention the references  \cite{ingrid, luciana, guilherme, januario, donizete}, with the first four using the Flesch Index of readability and the latter, The  Gunning fog Index. This is, of course, due to the absence of studies involving the adaptation of the foreign readability indexes to the Portuguese language.

This Article is organized as follows: in Section  \ref{indices-legibilidade} we present a brief review of the readability of interest indices, including their original formulas. In Section  \ref{algoritmos} we show the algorithms responsible for counting characters, words, sentences, and syllables of any text. The reasons characters/words, words/sentences and other combinations are the basic variables of the main known readability indexes. The adaptation of formulas for the Portuguese language is shown in Section  \ref{formulas-portugues}. The overview of the ALT program and its features are covered in Section \ref{alt}. In order to know the ALT’s degree of accuracy, in Section \ref{comparacoes} we compared the indices obtained by ALT with those obtained by the original formulas. Finally, we discussed the limitations of the application of readability formulas in Section  \ref{limitacoes} and concluded this Article in Section \ref{conclusoes}.


\section{The Readability Indexes -- original formulas}
\label{indices-legibilidade}

In this section we discuss the readability indexes used in the ALT program – textual readability analysis. 

\subsection{ \textit{Flesch reading ease} }
\label{legibilidade-de-flesch}

The Flesch Reading Index is one of the oldest methods capable of quantifying the ``difficulty understanding'' texts, according to the very words of its creator, Rudolf Flesch  \cite{Flesh}.The method was developed in 1943 and the formula was revised in 1948, being normalized in a scale ranging from 0 (minimum readability) to 100 (maximum readability). The formula, focused only on English texts, is given by

\begin{equation}
	\label{formula-flesch}
	\text{Flesch reading ease} = 206.835 - 1.015 \times \left( \dfrac{\text{words}}{\text{sentences}}\right) - 84.6 \times \left( \dfrac{\text{syllables  }}{\text{words  }}\right) .
\end{equation}

As Flesch's readability Index was presented to the public many decades before the popularization of personal computers, the use of the formula was impractical in long articles or books. In this sense, the author recommended the sampling criterion: the counting of syllables, words and sentences could be made from 100 words, three to five texts in articles and 25 to 30 texts in a book. The choice of texts could follow a certain pattern, such as starting from the third paragraph of each page, for example \cite{Flesh}.

With the advent of computer popularization, the counting of words and sentences in texts as long as those of books became a very simple process. But it must be pointed out that the syllable count still remains a challenge. There is currently no algorithm capable of providing, without errors, the amount of syllables of a word, and not even the syllable concept is a consensus among linguists \cite{theguardian}.

\subsection{ \textit{Gunning fog index} }
\label{gunning-fog-index}

The Gunning fog index , often translated literally as Índice de Nevoeiro de Gunning, was developed by Robert Gunning in 1952. Gunning, a consultant who came to work for United Press, the Wall Street Journal and Newsweek,  was the first to present a readability formula that estimates the years of formal education that a person must have in order to understand the text without difficulty  \cite{Gunning}. For example, a text with a 12-point index would be suitable for a reader with high school education degree, which revolves around 12 years of studies (3 years of high school + 9 years of elementary school). This scale, based on years of studies rather than the arbitrary centigrade scale, is now known as level of education or level of schooling (grade level ).

The formula for obtaining the level of text instruction is given by

\begin{equation}
	\text{Gunning fog index} = 0.4 \times \left( \dfrac{\text{words}}{\text{sentences}}\right) + 40 \times \left( \dfrac{\text{complex words}}{\text{words}}\right) .
\end{equation}

As with the Flesch Readability Index, the Gunning Fox Index is based on the concepts of “complex sentences” and “complex words.” Complex sentences, in the sense of Gunning, are those very long in terms of the amount of words: note that the first variable (words/sentences) is one of the determinants in the text level of instruction. It is intuitive that large sentences, composed of much recursion  \cite{Recursividade}, make reading difficult. It is not rare that even an educated reader has to reread a very long sentence in order to understand the information contained in it.

Over the complex words, Gunning defines them as being those that contain three or more syllables. Proper names, family jargon, and composite words should not be taken into account. It is interesting to note the difference between Flesch and Gunning: while the former encompasses a broad spectrum of words ranging from the very simple (monosyllabic), passing through the moderate (dissyllabic and trisyllabic) and reaching the very difficult (with 4 or more syllables), the second highlights that words are either common or complex. In this sense, the Gunning Fog Index considers that the words ``message'' and ``heterozygote'' have the same degree of complexity.

\subsection{ \textit{Automated readability index} }

The ARI -- Automated readability index -- was developed by E. A. Smith and R. J. Senter in 1967. As pointed out in the seminal paper  \cite{Smith}, its objective was to offer a readability index for books, reports and technical manuals of the United States Air Force with the purpose of decreasing the time of extracting information from these documents.

Until 1967, there were at least three methods of obtaining textual readability indices. Two of them were the Flesch Readability Index and the Gunning Fox Index which, as already pointed out, used the number of syllables to infer the complexity of a word. The other was the 1948 Dale-Chall algorithm  \cite{DaleChall}, which presented the index in a scale of 4.9 (or less) up to 9.9 points. The problem with the latter was that the formula depended on comparing the text with a list of 763 “very common words”, such as yes and  no. This index was quite interesting for children's texts, since the repertoire of words known to children is small. However, the formula was not suitable for texts aimed at the adult public. It was at this juncture that the authors of the ARI method presented a formula that avoided the complexity involved in the syllable count of the Flesch Readability Index and the presence of a list of words of Dale-Chall Index. The Automated Readability Index, which is based on the grade level  scale, is given by

\begin{equation}
	\label{formula-ari}
	\text{ARI} = - 21.43 + 0.50 \times \left( \dfrac{\text{words}}{\text{sentences}}\right)  + 4.71 \times \left( \dfrac{\text{characters}}{\text{words}}\right).
\end{equation}

\noindent As you can see, the Automated Readability Index is also based on the concept of ``complex words'' and ``complex sentences.'' The advantage of ARI is the ease with which these concepts can be quantified, since it is enough to count the number of strokes, words and sentences. The count of these three variables is quite simple today with the help of a word processor such as Microsoft Word and the like.

\subsection{ \textit{Flesch–Kincaid grade level} }

In 1975, J. Peter Kincaid and collaborators recalculated three readability indices (ARI, Gunning fog index and Flesch reading ease) for texts related to the United States Navy. In addition, the formula of Flesch rewritten on the grade level scale, now known as the Flesch-Kincaid grade level [(Nível de Instrução de Flesch-Kincaid), was also presented:

\begin{equation}
	\label{formula-flesch-kincaid}
	\text{Flesch–Kincaid grade level} = - 15.59 + 0.39 \times \left( \dfrac{\text{words  }}{\text{sentences}}\right) + 11.8 \times \left( \dfrac{\text{syllables  }}{\text{words  }}\right).
\end{equation}

The conversion formula between the centigrade scales (from zero to 100) and the level of instruction, obtained from the manipulation of the Equations  (\ref{formula-flesch}) and (\ref{formula-flesch-kincaid}), is given by

\begin{equation}
	\label{formula-conversao-escalas}
	\text{Flesch–Kincaid grade level} = 63.88 - 0.38424\times (\text{Flesch reading ease}) - 20.7 \times \left( \dfrac{\text{syllables}}{\text{words  }}\right).
\end{equation}

\subsection{ \textit{Coleman–Liau index} }

The Coleman–Liau index (Índice de Coleman-Liau] follows the criterion adopted by the ARI method, in the sense that it was developed with the purpose of being an index of easy computational implementation. Its formula, created by Meri Coleman and T. L. Liau, is given by  \cite{Coleman}

\begin{equation}
	\label{formula-coleman-liau}
	\text{Coleman–Liau grade level} = -15.8 - 2.96 \times \left( \dfrac{\text{sentences  }}{\text{words}}\right) + 5.88 \times \left( \dfrac{\text{letters}}{\text{words}}\right).
\end{equation}

\noindent This index is quite similar to ARI. The most notable difference is to infer the complexity of sentences by the sentence/word ratio, which is the inverse of what appears in the ARI and in the other indexes already presented (words/sentences).

\subsection{ \textit{Indice Gulpease} }

Developed by Gruppo Universitario Linguistico Pedagogico (GULP) of the University of Rome La Sapienza in 1987, the Indice Gulpease (Gulpease Index) provides a number for the readability of texts in Italian language. Also delimited by the centigrade scale, its formula is given by

\begin{equation}
	\label{formula-gulpease}
	\text{Indice Gulpease} = 89 + 300 \times \left( \dfrac{\text{sentences  }}{\text{words}}\right) - 10 \times \left( \dfrac{\text{letters}}{\text{words}}\right).
\end{equation}

This index also does not use the criterion of the number of syllables to delimit complex words.


\section{The algorithms}
\label{algoritmos}

\begin{figure}[H]
	\centering
	\includegraphics[width=0.4\linewidth]{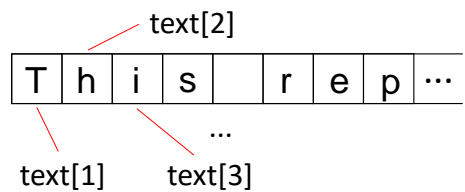}
	\caption{Storing the content of a document that starts with ``This report presents ...'' in the text vector. In this example, the first five vector components are the characters \texttt{T}, \texttt{h}, \texttt{i}, \texttt{s} and a blank space. An arbitrary k-th component of the vector is represented by  text[k].}
	\label{fig1}
\end{figure}

To count the number of characters, words, sentences, and syllables in the ALT software, the first procedure is to store all the characters in the document in a vector, which we will call text, N components, as example shown in Figure  \ref{fig1}. These characters can include letters, numbers, punctuation marks, other symbols arranged on the keyboard, and other symbols found in other languages. The count of these four mentioned variables is based on the analysis of the text vector components, whose algorithms are described in the sections below.

\subsection{Characters count}
\label{contagem-caracteres}

We consider all letters, both upper and lower case, and the - symbol (hyphen), as well as numbers, signs, and other symbols as characters. We then define the variable qntCharacters, which represents the total character of the text. As text vector components are read, a function is called to know if the symbol stored in the vector is a letter or hyphen. In either case, the qntCharacteristic variable is incremented in one unit.

At the end of the reading of the $N$ components of the text vector, the qntCharacter variable will contain the total text characters of the document.

\subsection{Words counting}

For words counting, the initial idea is contained in \cite{Smith}: incrementar em uma unidade a variável \texttt{qntPalavras} cada vez que um espaço vazio (produzido pela barra de espaço do teclado) é encontrado no vetor \texttt{texto}. Essa é uma solução simplificada, mas não inteiramente correta. Como os espaços estão entre as palavras, há, em geral, um espaço vazio a menos do que o total de palavras de uma sentença. E devemos considerar também o caso dos parágrafos. Três parágrafos com apenas uma palavra cada não contêm nenhum espaço vazio.

The algorithm integrated into the ALT is then given by the following statement: Increment the variable qntWords  whenever the text[k] component is the last vector, or an empty space (`` ''), or a carriage return, produced by the Enter/Return key (``$\backslash$n''), at the same time as the previous component (text[k-1]) is neither the previous objects nor the hyphen symbol.

\subsection{Sentences counting}

To count the sentences, we increase the variable qntSentences whenever the text[k] component is equal to a period (.), an exclamation point (!), or a question mark (?), or a semicolon (;) while text[k-1] is not any of these punctuation marks, in order not to count further sentences in cases where these punctuation marks appear in sequence. The semicolon (;) was considered here by the frequency in which they are displayed to quote lists of items, ideas, separation of coordinated prayers, and other cases, which would represent a long sentence, interfering with the final index obtained.

\subsection{Syllables counting}

The starting point is a peculiarity of the Portuguese language: the total of syllables of a word is equal to the number of vowels contained in it. The problem is then to eliminate from the counting all the I and u semivowels that can appear in the diphthongs and triphthongs. This is not a simple task, and what we present here is just an algorithm that returns the approximate number of syllables of a word.

The first stage is to store all known vowels, diphthongs and triphthongs, including the presence of eventual accents, in vowel, diphthong and triphthongs vectors:

\begin{algorithm}[ht]
	vowel  = [a, ã, â, á, à, e, é, ê, i, í, o, ô, õ, ó, u, ú]
	
	diphthong  = [ ãe, ai, ão, au, ei, eu, éu, ia, ie, io, iu, õe, oi, ói, ou, ua, ue, uê, ui ]
	
	triphthong  = [uai, uei, uão, uõe, uiu, uou]
\end{algorithm}

Then we increment the variable qntSyllables every time text[k] is equal to any of the components of the vowel vector or its corresponding in capital letters. The last step is to disregard the semivowels of diphthongs and triphthongs.

For the case of diphthongs, we start by reducing the variable qntSyllables  in one unit every time text[k-1] + text[k] is equal to any of the diphthong vector components while text[k-2] is a consonant.

To eliminate the triphthongs' semivowels, we reduced the qntSyllables variable to one unit each time text[k-2] + text[k-1] + text[k] equals any of the triphthong vector components.

The procedure described in this algorithm is able to correctly return the number of syllables of approximately 96\% of the words.

\section{Adaptation of the formulas to Portuguese Language}
\label{formulas-portugues}

First of all, it is interesting to observe the six readability formulas used in ALT software, equations (1)-(6). All of them depend linearly on two variables, so that they can be represented generically by

\begin{equation}
	\label{plano}
	f(x,y) = C_1 + C_2 \;x + C_3 \;y.
\end{equation}

\noindent The variables are being represented by x and y, which can be words/sentences or syllables/words, for example, and Ck is the k-th coefficient. This means that, from a multiple linear regression, it is possible to obtain the plane given by the equation  (\ref{plano}) that best fits a certain set of points. At this point, it is worth emphasizing why we do not use other readability formulas in ALT software, such as  G. Harry McLaughlin's SMOG Grade \cite{Smog},

\begin{equation}
	\label{smog}
	\text{SMOG grade} = 1.0430 \sqrt{ 30\times \left( \dfrac{\text{polysyllables}}{ \text{sentences} }\right)  } + 3.1291,
\end{equation}

\noindent where polysyllables in this case, corresponds to the number of words with three or more syllables. A non-linear dependence between the level of instruction and the polysyllables/sentences variable makes the surface adjustment  given by the Equation  (\ref{smog}) more difficult: while only three points determine a plane, infinite points are needed to characterize any other surface.

To adapt to Portuguese Languages the formulas given by equations  (\ref{formula-flesch})-(\ref{formula-gulpease}) means changing the  C k coefficients so that the legibility indices for documents in that language are as close as possible to those obtained in the English (or Italian) translation, in the case of the Gulpease index of the document  from the formulae with the original coefficients. This could be done from a multiple linear regression using the following procedure:

\begin{enumerate}
	\item We selected a sample with $N = 100$ texts of several genres in the Portuguese language \cite{ListaTextos}.
	
	\item Using the algorithms of Section  \ref{algoritmos} written in JavaScript, we calculated all the variables of interest of the $N$ texts: $\left\lbrace x_1, x_2, \ldots, x_N\right\rbrace $ e $\left\lbrace y_1, y_2, \ldots, y_N\right\rbrace $.
	
	\item We obtained the English (or Italian) versions of the $N$ texts.
	
	\item4.	We obtained the readability indexes of the translated versions of the texts (GLk, k-nth grade level) using tools available on the web, as the Readability Test Tool  \textit{Readability Test Tool} \cite{RTT} and the Farfalla Project \cite{Darian}.
	
	\item From the previous data, we assembled an  N 3 matrix, the first column consisting of $x_k$ values, the second by the values of $y_k$, both obtained in step 2, and the third column, the indices of the texts translated versions, obtained in step 4.
	
	\item We obtained $C_1$, $C_2$ and $C_3$  coefficients through multiple linear regression of the matrix data formed in step 5.
\end{enumerate}

In the following sections we present the results obtained.

\subsection{Flesch Reading Ease score}
\label{indices-portugues}

The results obtained are summarized in Table 1 and Figure 2.

\begin{table}[ht]
	\centering
	\caption{Coefficients for the Flesch Readability Index obtained by multiple linear regression.}
	\begin{tabular}{|lllll|}
		\hline
		\multicolumn{5}{|c|}{$R^2 = 0,890742$}                                                                                                               \\ \hline
		\multicolumn{1}{|l|}{}      & \multicolumn{1}{l|}{Reference}         & \multicolumn{1}{l|}{Value}      & \multicolumn{1}{l|}{Standard error} & p-value \\ \hline
		\multicolumn{1}{|l|}{$C_1$} & \multicolumn{1}{l|}{\textit{Intercept}} & \multicolumn{1}{l|}{226.614882} & \multicolumn{1}{l|}{8.744455}    & 0.00000 \\ \hline
		\multicolumn{1}{|l|}{$C_2$} & \multicolumn{1}{l|}{words/sentences} & \multicolumn{1}{l|}{$-1.036134$}  & \multicolumn{1}{l|}{0.0930814}   & 0.00000 \\ \hline
		\multicolumn{1}{|l|}{$C_3$} & \multicolumn{1}{l|}{syllables/words}   & \multicolumn{1}{l|}{$-72.451284$} & \multicolumn{1}{l|}{4.336399}    & 0.00000 \\ \hline
	\end{tabular}
\end{table}

\begin{figure}[ht]
	\centering
	\includegraphics[width=0.95\linewidth]{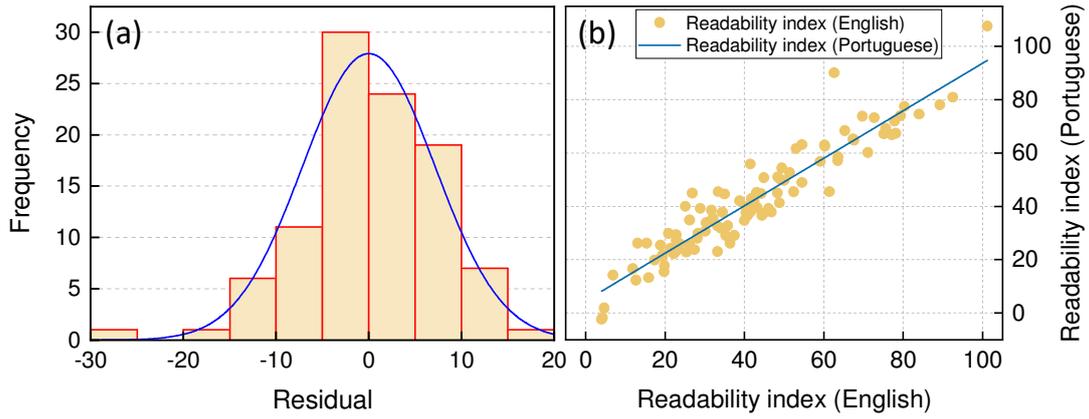}
	\caption{Result of multiple linear regression for Flesch Readability Index. (a) Histogram of the difference between the readability index obtained (for Portuguese texts using ALT software) and the predicted value (same English text from RTT tool). The blue curve represents the adjustment of the normal distribution around the histogram data. (b) the graph of dependence between the obtained and predicted readability indices.}
	\label{fig2}
\end{figure}

Formula considered for the Portuguese language:

\begin{equation}
	\label{flesch-pt}
	\text{Flesch Reading Ease score} = 227 - 1{,}04 \times \left( \dfrac{\text{words}}{\text{sentences}}\right) - 72 \times \left( \dfrac{\text{syllables}}{\text{words}}\right) .
\end{equation}

Comentários: If the formula (\ref{flesch-pt}) is reapplied to the 100 texts in the sample, 84 of them [see Fig. \ref{fig2}(a)] will present a difference in the readability index less than or equal to 10 points in relation to the index obtained with the original formula applied in the English version of the text. The blue fit curve obtained from the data in Fig. \ref{fig2}(b) shows that the variables $x$ and $y$ explain 89.1\% of the variance of the readability index.

\subsection{ Gunning Fog Index}

Our adaptation to the Gunning Fog Index makes use of a different definition of “complex word”. Whereas the Gunning fog index considers complex words to be those that contain three or more syllables, as pointed out in Section  \ref{gunning-fog-index}, here we do a verification of the word with a Linguateca word databases, a resource center for computational processing of the Portuguese language  \cite{Linguateca}. This modification was made in order to obtain a different method in the calculation of the readability of a text, since another index, such as Flesch-Kincaid, also uses the syllable counting criterion to define what is  a complex word or not.

The used word databases is available in one of the Linguateca's URLs  \cite{banco-palavras}, in the ``All Brazilian bodies'' option, column ``List of total frequency of forms in the body''. We consider only the first five thousand items to form the ALT program databases. The criterion used was then to consider as ``complex word'' all those beyond this number.

The results obtained are summarized in Table 2 and Figure 3.

\begin{table}[ht]
	\centering
	\caption{Coefficients for the Gunning Fog Readability Index obtained by multiple linear regression.}
	\begin{tabular}{|lllll|}
		\hline
		\multicolumn{5}{|c|}{$R^2 = 0,77333$}                                                                                                                       \\ \hline
		\multicolumn{1}{|l|}{}      & \multicolumn{1}{l|}{Reference}                  & \multicolumn{1}{l|}{Value}    & \multicolumn{1}{l|}{Standard error} & p-value \\ \hline
		\multicolumn{1}{|l|}{$C_1$} & \multicolumn{1}{l|}{\textit{Intercept}}          & \multicolumn{1}{l|}{1.00156}  & \multicolumn{1}{l|}{1.28036}     & 0.43599 \\ \hline
		\multicolumn{1}{|l|}{$C_2$} & \multicolumn{1}{l|}{words/sentences}          & \multicolumn{1}{l|}{0.49261}  & \multicolumn{1}{l|}{0.02764}     & 0.00000 \\ \hline
		\multicolumn{1}{|l|}{$C_3$} & \multicolumn{1}{l|}{complex words/words} & \multicolumn{1}{l|}{18.66057} & \multicolumn{1}{l|}{5.6943}      & 0.00146 \\ \hline
	\end{tabular}
\end{table}

\begin{figure}[ht]
	\centering
	\includegraphics[width=0.95\linewidth]{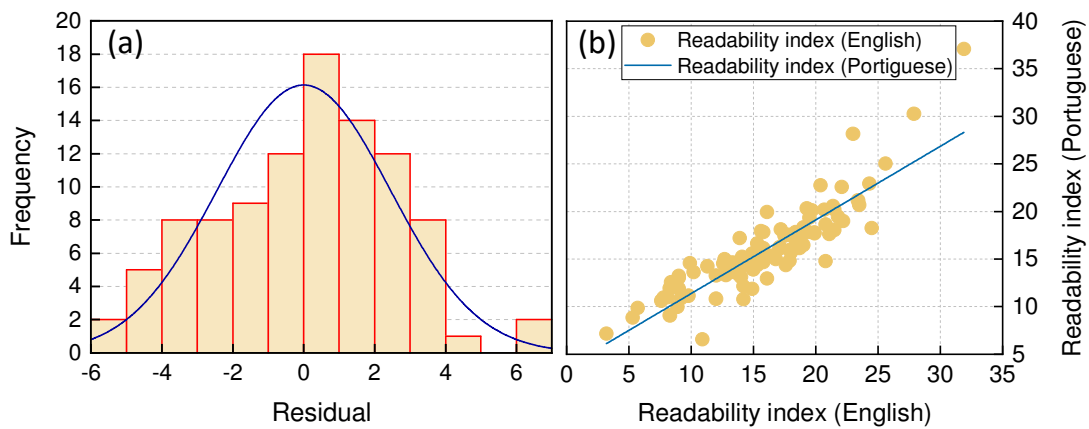}
	\caption{Result of multiple linear regression for the Gunning Fog Index. (a)-(b): Idem Figure  \ref{fig2}.}
	\label{fig3}
\end{figure}

Formula considered for the Portuguese language:

\begin{equation}
	\label{gunning-fog-pt}
	\text{Gunning fog index } = 0.49 \times \left( \dfrac{\text{words}}{\text{sentences}}\right) + 19 \times \left( \dfrac{\text{complex words}}{\text{words}}\right)
\end{equation}

Comments: 

\begin{enumerate}
	\item The standard error in the $C_1$  coefficient (relative to intercept) is greater than its own value. In other words, we can consider $C_1 = 0$ for all purposes. This is corroborated by the high p-value, much higher than the standard normally adopted of 5\%. In fact, the original formula does not have this coefficient, which indicates a good adaptation for the Portuguese language at this point.
	
	\item If the equation  (\ref{gunning-fog-pt}) ) is applied again to the 100 sample texts, 73 of them  [see Fig. \ref{fig3}(a)] shall show a difference in the readability index less than or equal to 3 points in relation to the index obtained with the original formula applied in the English version of the text. This is not a result that we can find interesting. On average, this means that 27\% of the texts analyzed will have a difference in the index of more than 3 points. Most likely this comes from the great uncertainty in the $C_3$ coefficient , whose standard error is almost a third of its value. This coefficient is linked to the variable complex words/words. This can be an indication that an update in the word databases  may be required.
	
	\item The blue adjustment curve obtained from the data in Figure \ref{fig3}(b) shows that the  variables x and y explain 77.3\% of the variance of the Gunning index. The great uncertainty in the $C_3$ coefficient explains this relatively low value for $R^2$.
\end{enumerate}

\subsection{ Automated Readability Index }

The results obtained are summarized in Table 2 and Figure 3.

\begin{table}[ht]
	\centering
	\caption{Coefficients for the ARI obtained by multiple linear regression.}
	\begin{tabular}{|lllll|}
		\hline
		\multicolumn{5}{|c|}{$R^2 = 0.93696$}                                                                                                                          \\ \hline
		\multicolumn{1}{|l|}{}      & \multicolumn{1}{l|}{Reference}                  & \multicolumn{1}{l|}{Value}       & \multicolumn{1}{l|}{Standard error} & p-value \\ \hline
		\multicolumn{1}{|l|}{$C_1$} & \multicolumn{1}{l|}{\textit{Intercept}}          & \multicolumn{1}{l|}{$-$20.26065} & \multicolumn{1}{l|}{1.67994}     & 0.00000 \\ \hline
		\multicolumn{1}{|l|}{$C_2$} & \multicolumn{1}{l|}{letters/words}          & \multicolumn{1}{l|}{4.57058}     & \multicolumn{1}{l|}{0.36508}     & 0.00000 \\ \hline
		\multicolumn{1}{|l|}{$C_3$} & \multicolumn{1}{l|}{words/sentences} & \multicolumn{1}{l|}{0.43664}     & \multicolumn{1}{l|}{0.01834}     & 0.00000 \\ \hline
	\end{tabular}
\end{table}

\begin{figure}[ht]
	\centering
	\includegraphics[width=0.95\linewidth]{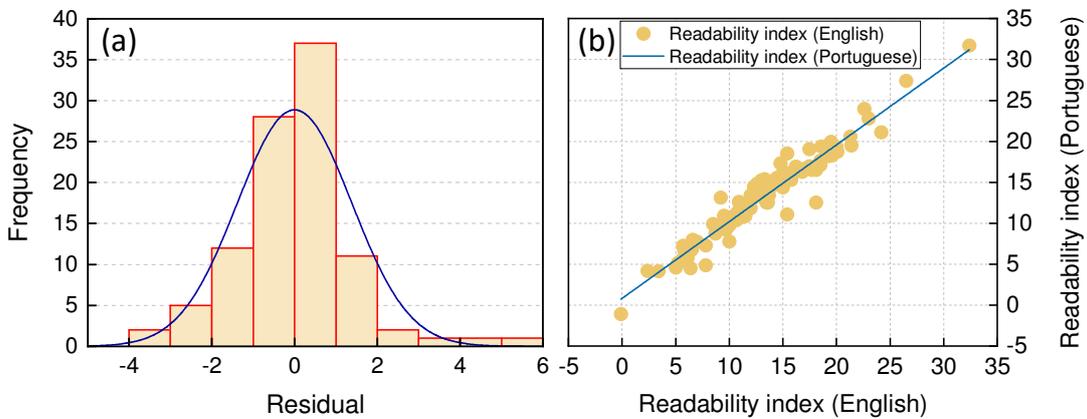}
	\caption{Result of multiple linear regression for ARI. (a)-(b): Idem Figure  \ref{fig2}.}
	\label{fig4}
\end{figure}

Formula considered for the Portuguese language:

\begin{equation}
	\label{ari-pt}
	\text{Automated Readability Index} = 0.44 \times \left( \dfrac{\text{words}}{\text{sentences}}\right)  + 4.6 \times \left( \dfrac{\text{characters}}{\text{words}}\right) - 20.
\end{equation}

Comments: 

\begin{enumerate}
	\item If the equation  (\ref{ari-pt}) is applied again to the 100 sample texts, 88 of them  [see Fig. \ref{fig4}(a)] shall show a difference in the readability index less than or equal to 2 points in relation to the index obtained with the original formula applied in the English version of the text, which represents an excellent result. These facts corroborate with the relatively low standard errors in the three coefficients and the optimal adjustment observed in the points of Figure \ref{fig4}(b).
\end{enumerate}

\subsection{ Flesch-Kincaid Grade Level }

The results obtained are summarized in Table 4 and Figure 5.

\begin{table}[ht]
	\centering
	\caption{Coefficients for the   Flesch-Kincaid Grade Level  obtained by multiple linear regression.}
	\begin{tabular}{|lllll|}
		\hline
		\multicolumn{5}{|c|}{$R^2 = 0.92273$}                                                                                                                 \\ \hline
		\multicolumn{1}{|l|}{}      & \multicolumn{1}{l|}{Reference}         & \multicolumn{1}{l|}{Value}       & \multicolumn{1}{l|}{Standard error} & p-value \\ \hline
		\multicolumn{1}{|l|}{$C_1$} & \multicolumn{1}{l|}{\textit{Intercept}} & \multicolumn{1}{l|}{$-18.11589$} & \multicolumn{1}{l|}{1.6077}      & 0.00000 \\ \hline
		\multicolumn{1}{|l|}{$C_2$} & \multicolumn{1}{l|}{words/sentences} & \multicolumn{1}{l|}{0.36001}     & \multicolumn{1}{l|}{0.01712}     & 0.00000 \\ \hline
		\multicolumn{1}{|l|}{$C_3$} & \multicolumn{1}{l|}{syllables/words}   & \multicolumn{1}{l|}{10.35177}    & \multicolumn{1}{l|}{0.79701}     & 0.00000 \\ \hline
	\end{tabular}
\end{table}

\begin{figure}[ht]
	\centering
	\includegraphics[width=0.95\linewidth]{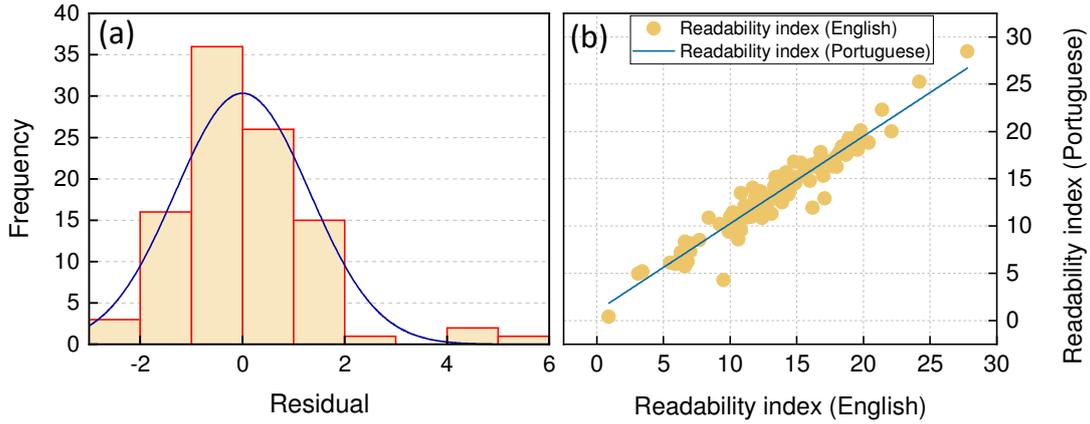}
	\caption{Result of multiple linear regression for the  Flesch-Kincaid Grade Level. (a)-(b): Idem Figure \ref{fig2}.}
	\label{fig5}
\end{figure}

Formula considered for the Portuguese language:

\begin{equation}
	\label{flesch-kincaid-pt}
	\text{Flesch-Kincaid Grade Level  } = 0.36 \times \left( \dfrac{\text{words}}{\text{sentences}}\right) + 10.4 \times \left( \dfrac{\text{syllables}}{\text{words}}\right) - 18.
\end{equation}

Comments: 

\begin{enumerate}
	\item 1.	As in  ARI, we have here an excellent adaptation indicated by the low standard errors in the coefficients together with a high value of $R^2$, as it can be observed by the results presented in figures  \ref{fig5}(a) and \ref{fig5}(b).
\end{enumerate}

\subsection{ Coleman–Liau index }

The results obtained are summarized in Table 5 and Figure 6.

\begin{table}[ht]
	\centering
	\caption{Coefficients for the Coleman-Liau Index obtained by multiple linear regression.}
	\begin{tabular}{|lllll|}
		\hline
		\multicolumn{5}{|c|}{$R^2 = 0.89221$}                                                                                                                \\ \hline
		\multicolumn{1}{|l|}{}      & \multicolumn{1}{l|}{Reference}         & \multicolumn{1}{l|}{Value}     & \multicolumn{1}{l|}{Standard error} & p-value \\ \hline
		\multicolumn{1}{|l|}{$C_1$} & \multicolumn{1}{l|}{\textit{Intercept}} & \multicolumn{1}{l|}{$-13.66302$} & \multicolumn{1}{l|}{1.61422}     & 0.00000 \\ \hline
		\multicolumn{1}{|l|}{$C_2$} & \multicolumn{1}{l|}{letters/words}    & \multicolumn{1}{l|}{5.39801}   & \multicolumn{1}{l|}{0.27242}     & 0.00000 \\ \hline
		\multicolumn{1}{|l|}{$C_3$} & \multicolumn{1}{l|}{syllables/words}   & \multicolumn{1}{l|}{$-20.57984$}  & \multicolumn{1}{l|}{6.67523}     & 0.00000 \\ \hline
	\end{tabular}
\end{table}

\begin{figure}[ht]
	\centering
	\includegraphics[width=0.95\linewidth]{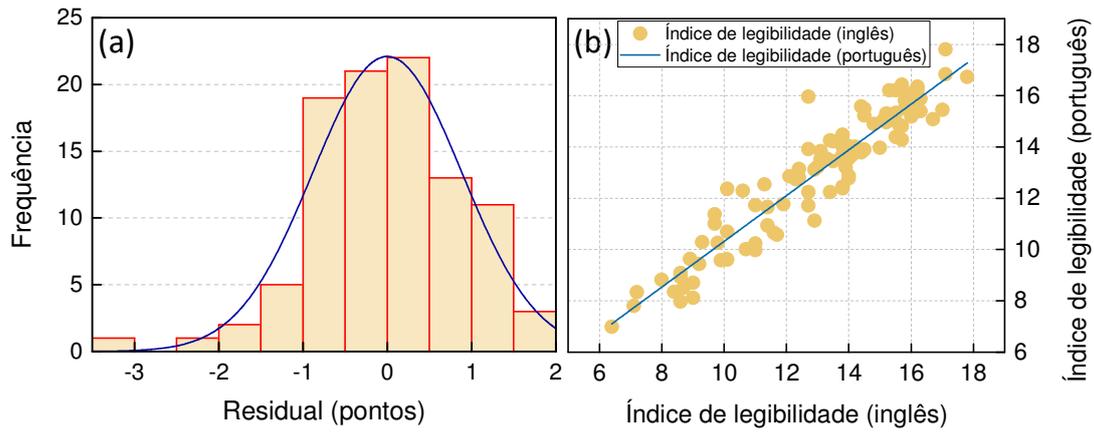}
	\caption{Result of multiple linear regression for the  Coleman-Liau Index. (a)-(b): Idem Figure  \ref{fig2}.}
	\label{fig6}
\end{figure}

Formula considered for the Portuguese language:

\begin{equation}
	\label{coleman-liau-pt}
	\text{Coleman-Liau index } = 5.4 \times \left( \dfrac{\text{characters}}{\text{words}}\right) - 21 \times \left( \dfrac{\text{sentences}}{\text{words}}\right) - 14.
\end{equation}

Comments: 

\begin{enumerate}
	\item 1.	As in  ARI, we have here an excellent adaptation indicated by the low standard errors in the coefficients together with a high value of $R^2$, as it can be observed by the results presented in figures  \ref{fig6}(a) and \ref{fig6}(b).
\end{enumerate}

\subsection{ Gulpease index}

The Gulpease  Index did not show, within the limits of the error margins, variations in its coefficients when we applied the procedure outlined in Section  \ref{formulas-portugues}. This is an indication that this index, in its original formula for Italian, can be used for texts in Portuguese. One reason for this may be the same Latin origin as these two languages.

\section{The ALT software}
\label{alt}

ALT -- Textual Readability Analysis -- is a computer program capable of returning, in quantitative terms, the level of ease of reading texts. It was developed by professors Marco Polo Moreno de Souza and Gleice Carvalho de Lima Moreno with the collaboration of professors Nelson Hein and Adriana Kroenke Hein written in JavaScript language. Through the algorithms described in Section  \ref{algoritmos} and the readability indices adapted for the Portuguese language, Equation (\ref{flesch-pt}) yo (\ref{coleman-liau-pt}), ALT provides the final result of the readability of a text by the arithmetic mean of four indexes operating on the instruction level scale:

\begin{equation}
	\label{resultado}
	\text{Result} = \dfrac{1}{4} \left( \text{FK} + \text{GF} + \text{ARI} + \text{CL} \right),
\end{equation}

\noindent where:

\begin{itemize}
	\item FK $=$ Flesch-Kincaid Grade Level,
	
	\item GF $=$ Gunning fog index,
	
	\item ARI $=$ Automated Readability Index and
	
	\item CL $=$ Coleman-Liau index.
\end{itemize}

\noindent We did not use for the final result the indices that operate on the centigrade scale because a conversion of scales would be necessary. However, the ALT program also provides individual metrics for the Flesch Readability Index and Gulpease Index, as they may be of interest to the user.

\subsection{General Vision}

\begin{figure}[ht]
	\centering
	\includegraphics[width=0.99\linewidth]{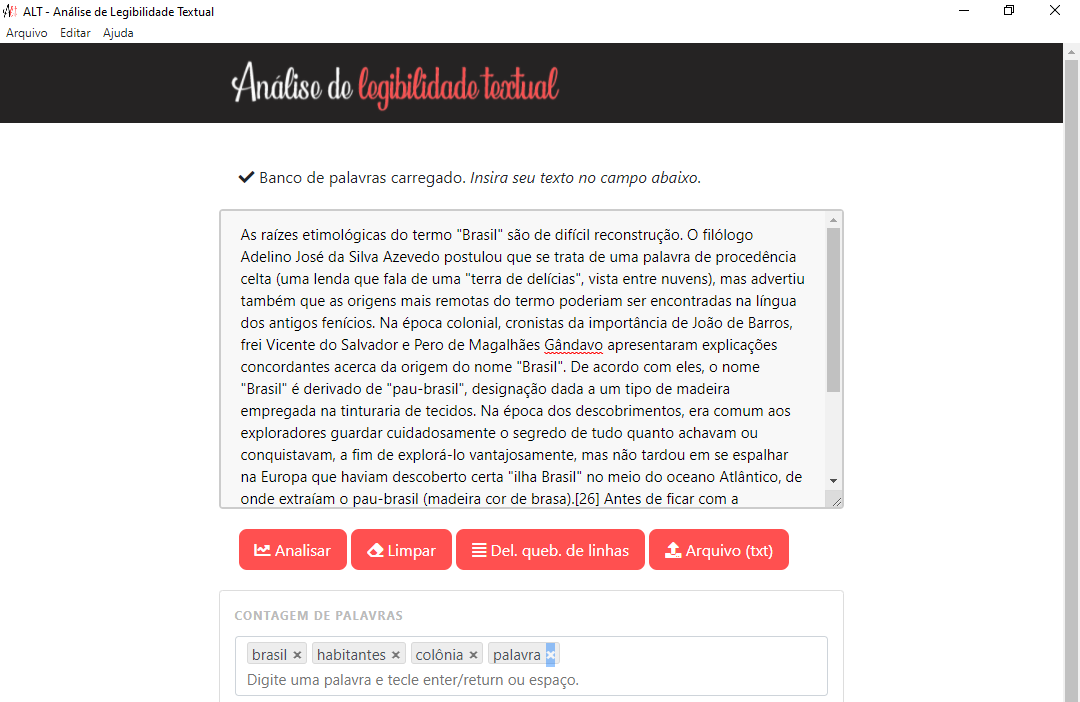}
	\caption{Part of the ALT program layout in version 1.1.0 for Windows.}
	\label{fig7}
\end{figure}

We introduce in Figure  \ref{fig7} the layout of the ALT program with the insertion of the Etymology section of the Wikipedia Brazil Article  \cite{brasil}. The readability indexes are obtained by clicking the analyze button, the results of which we show in the next subsection.

\subsection{The metrics}

\begin{figure}[ht]
	\centering
	\includegraphics[width=0.99\linewidth]{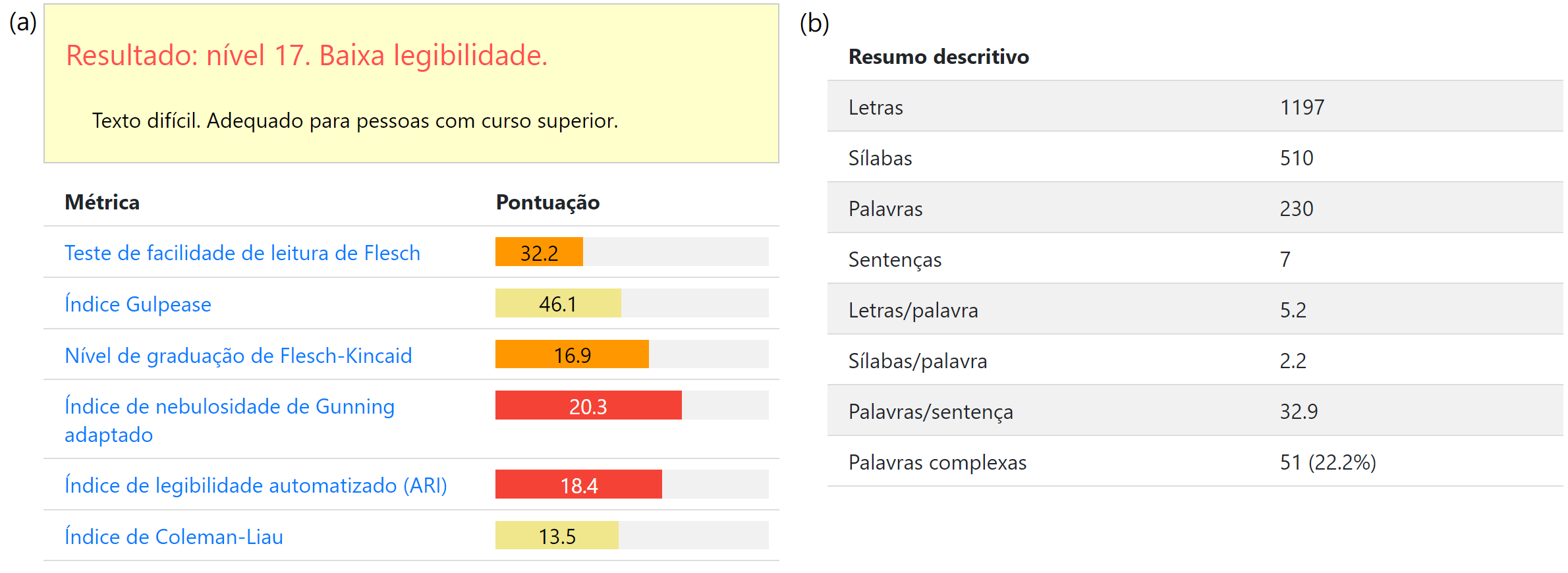}
	\caption{(a) Final result and specific legibility indices. (b) descriptive summary containing the variables of the text analyzed.}
	\label{fig8}
\end{figure}

The text readability is given in a yellow field as it can be seen in Figure  \ref{fig8}(a).The level of readability obtained by means of Equation (\ref{resultado}), is a number in general from 5 to 20. In addition, we also presented the readability in three degrees: low, medium and high readability, obtained by the following way:

\begin{itemize}
	\item Result below 13 points: high readability
	
	\item Result from 13 and below 17 points: medium readability.
	
	\item Result equal to or greater than 17 points: low readability.
\end{itemize}

Just below the result, the program shows the individual indexes captured using the six metrics: Flesch reading ease test, Gulpease Index, Flesch-Kincaid Grade Level, Gunning Fog Index, Automated Readability Index (ARI) and Coleman-Liau Index.

Finally, the program shows all variables of interest, as it can be seen in Figure  \ref{fig8}(b): Number of letters, syllables, words, sentences, and complex words, and also some of their ratios: Letters/word, syllables/word, and words/sentence.

\subsection{Search for specific words and the  word cloud}

\begin{figure}[ht]
	\centering
	\includegraphics[width=0.99\linewidth]{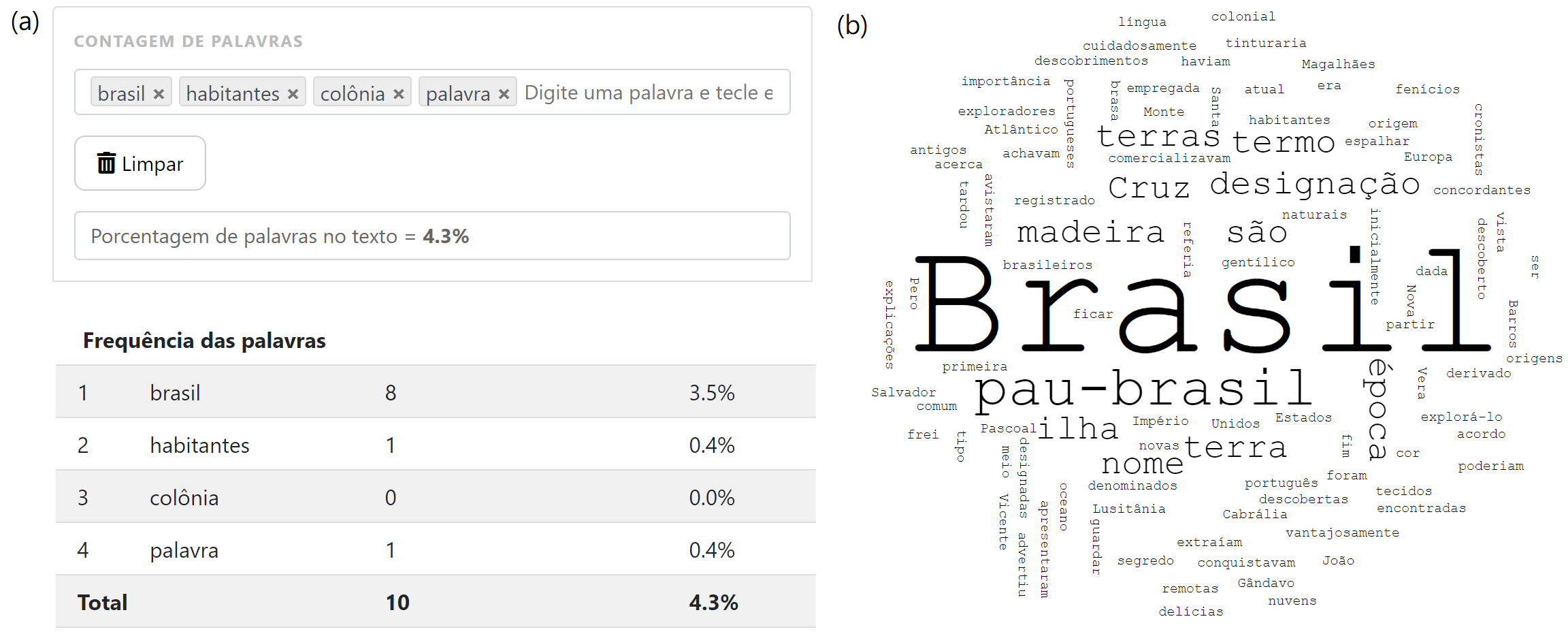}
	\caption{(a) Specific words counting. (b) Cloud of words.}
	\label{fig9}
\end{figure}

Another part of the ALT program is dedicated to analyzing the content of the text in terms of words and frequencies. Just below the text entry field, figures  \ref{fig7}, you can search for specific words within the text. Since the analyze button is clicked, the absolute and relative frequencies are shown in a table, as shown in figure \ref{fig9}(a).

The subject of the text can be inferred through a word cloud, where an image with words arranged in the horizontal or vertical directions is presented in sizes proportional to its frequencies in the text. This is a form of visual content analysis, where the content of the text can be quickly obtained through the greatest words in the cloud. In the example text shown, whose word cloud is shown in Figure  \ref{fig9}(b), we can clearly observe the prominence of the terms ``Brasil'', ``pau-brasil'', ``madeira'', ``Cruz'', ``termo'', among others. With this information, it is possible to know what the text is about, even without having read it:  a discussion about the origin of the name Brazil country.

It is important to point out that functional words, which have little role in the transmission of semantic information, have been removed from the cloud and may or may not be disregarded from word counting, which includes prepositions, articles, pronouns and conjunctions. They are listed in Appendix  \ref{lista-palavras-removidas}.

\subsection{Improvements suggestions}

Finally, the ALT program indicates in a text field some points that contribute to text with low readability. The sentences considered long (which we consider to be those formed by 30 to 45 words) and very long (formed by more than 45 words) are  highlighted in yellow and red, respectively. It is suggested for the user to consider dividing the long sentences into two and the very long sentences into two or more.

\begin{figure}[ht]
	\centering
	\includegraphics[width=0.75\linewidth]{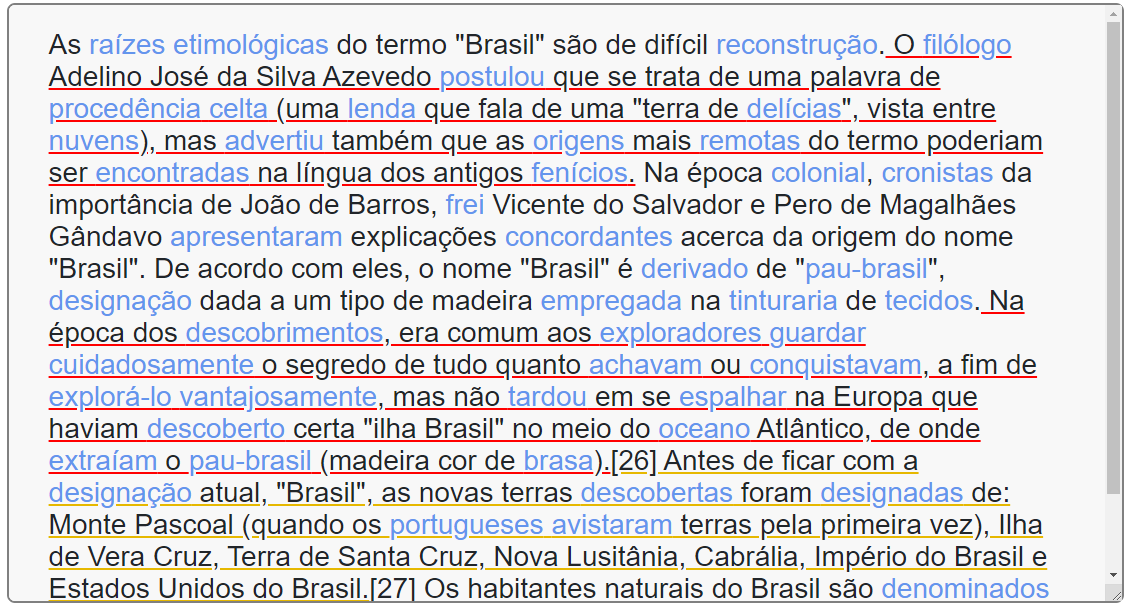}
	\caption{Notes highlighted by the ALT program. The sentences considered long or too long are highlighted in yellow and red, respectively. The ``complex'' words  are indicated by the light blue font.}
	\label{fig10}
\end{figure}

Whereas the  words considered ``complex'' are highlighted by light blue. It is suggested to replace them with simpler words at the discretion of each user.

\subsection{Extensions}

Currently, ALT software is available in seven extensions, one of which must be accessed online while the other six must be downloaded and used on the user's device. The latter are

\begin{enumerate}
	\item \texttt{exe}: for installation on Windows 64-bit devices.
	
	\item \texttt{apk}: for installation on Android-powered devices.
	
	\item \texttt{dmg}: For installation on MacOS-enabled devices.
	
	\item \texttt{deb}: For installation on Linux devices (some distributions).
	
	\item \texttt{AppImage}: for use (no installation required) on devices with Linux (some distributions).
	
	\item \texttt{Play Store}: available from the Play Store app store for Android devices.
\end{enumerate}

The source code of the ALT software consists of JavaScript, HTMAll seven extensions are produced from minor adaptations of the same project hosted on the GitHub platform. For the development of the extensions mainly aimed at desktops (exe, dmg, deb and AppImage), we used Electron, a framework connected to Node.js and the Chromium browser. Whereas for the development of extensions targeted to mobile devices, we used Android Studio together with its WebView component.

\begin{figure}[ht]
	\centering
	\includegraphics[width=0.45\linewidth]{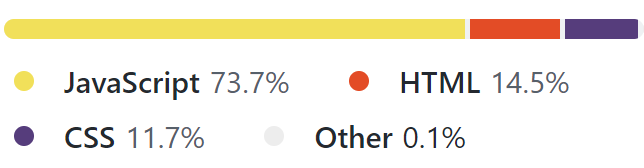}
	\caption{Composition of languages in the source code of the ALT program according to GitHub.}
	\label{fig11}
\end{figure}

\section{Comparing the formulas adapted with their original versions in English Language}
\label{comparacoes}

\begin{figure}[ht]
	\centering
	\includegraphics[width=0.99\linewidth]{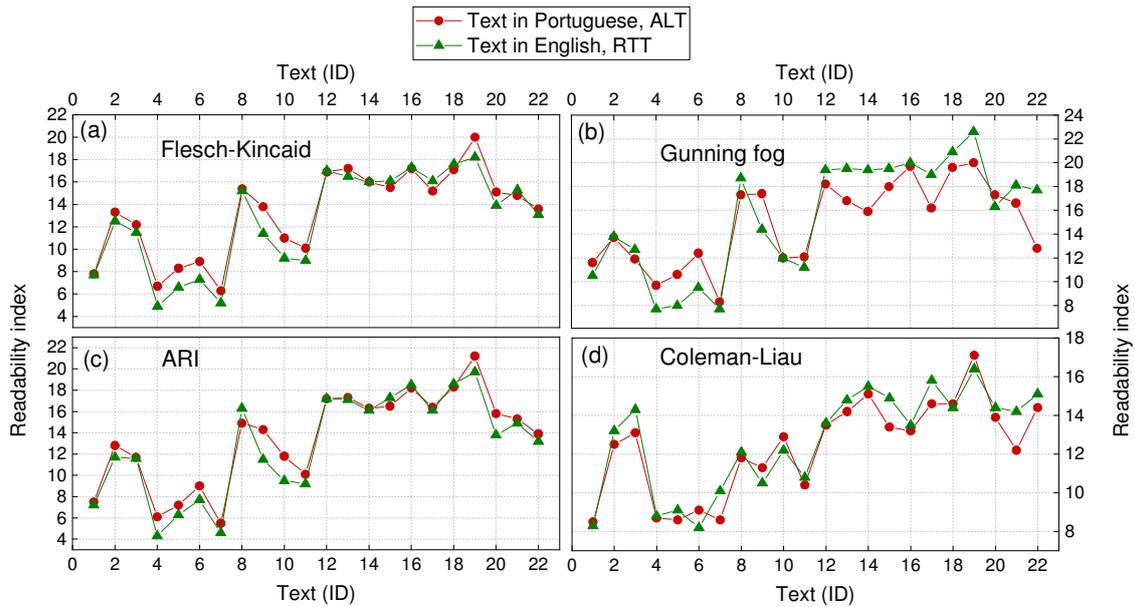}
	\caption{Comparison between the readability indexes of the texts of Appendix  \ref{tabela-textos-2}, represented by their ID.  Red circles represent the indices obtained by the ALT program, while green triangles indicate the indices obtained from the English versions of the texts, via the RTT tool.}
	\label{fig12}
\end{figure}

In this Section, we presented a comparative analysis between the results obtained by the ALT program, which makes use of the readability formulas adapted for the Portuguese language  [equações (\ref{flesch-pt}) to (\ref{coleman-liau-pt})], and those obtained from the Readability Test Tool (RTT) \cite{RTT}, which uses the original formulas to measure the English texts readability. The sample consists of 22 texts listed in Appendix  \ref{tabela-textos-2}. The comparative charts are shown in Figure  \ref{fig12}, where red circles indicate the readability indices obtained by the ALT program, while green triangles represent the indices obtained from the RTT tool. Clearly, we observed a strong correlation between the results obtained, which is corroborated by the data in table  \ref{correlacao}, where we have Pearson's correlation and the average difference between the results obtained from the two programs. The ``margin of error'' of the average difference was considered with two standard deviations for each side, which should cover around 95\% of the texts of any sample.

The results of Figure  \ref{fig12} and the Table  \ref{correlacao} reveal a lower quality in the adaptation of the Gunning fog  index, even if a high correlation $(> 0.9)$ was obtained. Although the mean difference was only 0.4, the data dispersion  is more than double that of any of the average differences obtained for the other indices. This indicates that an improvement in the adaptation of the Gunning fog index  needs to be made for a more robust match.

\begin{table}[H]
	\centering
	\caption{Correlation and average difference between ALT program and RTT tool results.}
	\label{correlacao}
	\begin{tabular}{|l|c|c|}
		\hline
		Index          & Correlation & Average difference   \\ \hline
		Flesch-Kincaid  & 98.0\%     & 0.7 $\pm$ 1.8     \\ \hline
		Gunning fog     & 91.3\%     & $-$ 0.4 $\pm$ 4.2 \\ \hline
		ARI             & 97.9\%     & 0.7 $\pm$ 2.0     \\ \hline
		Coleman-Liau    & 95.3\%     & $-0.4$ $\pm$ 1.6  \\ \hline
		Final result & 97.2\%     & 0.6 $\pm$ 2.0     \\ \hline
	\end{tabular}
\end{table}

\newpage
\section{Limitations of the textual readability formulae}
\label{limitacoes}

The readability indexes need to be used in quite a criterion, since not always a low index (on the 0-20 scale) indicates an easy reading text. Note, for example, the first propositions (from 1 to 2,013) contained in the book  Tractatus Logico-Philosophicus, by  Ludwig Wittgenstein (Appendix  \ref{tractatus}).

As it can be noted, this is a philosophical text that consists of a set of short sentences grouped within numbered propositions, which guarantees readability indexes below 8 points in the four metrics of the Grade Level criterion:

\begin{itemize}
	\item Flesch-Kincaid: 5.5
	
	\item Gunning fog: 6.6
	
	\item ARI: 4.3
	
	\item Coleman-Liau: 7.2,
\end{itemize}

\noindent with final result of 6 points. This would indicate, at first, that it would be a suitable text for children aged 11 to 12. However, this work is considered a very difficult reading text, even for specialists  \cite{marques}. This apparent incongruence arises because the variables that require the text are close to those found in children's texts: the variables letters/word, syllables/word and words/sentence are 4.3, 1.9 and 10.3, respectively, and there are only 8\% of complex words. The  complexity of Tractatus lies, among other points, in the need for deep reflection and much familiarity with concepts of modern logic by the reader. This is a good example of a text that has high readability at the same time as it can be well understood by only a few people outside the specialized academic environment.

\begin{figure}[H]
	\centering
	\includegraphics[width=0.99\linewidth]{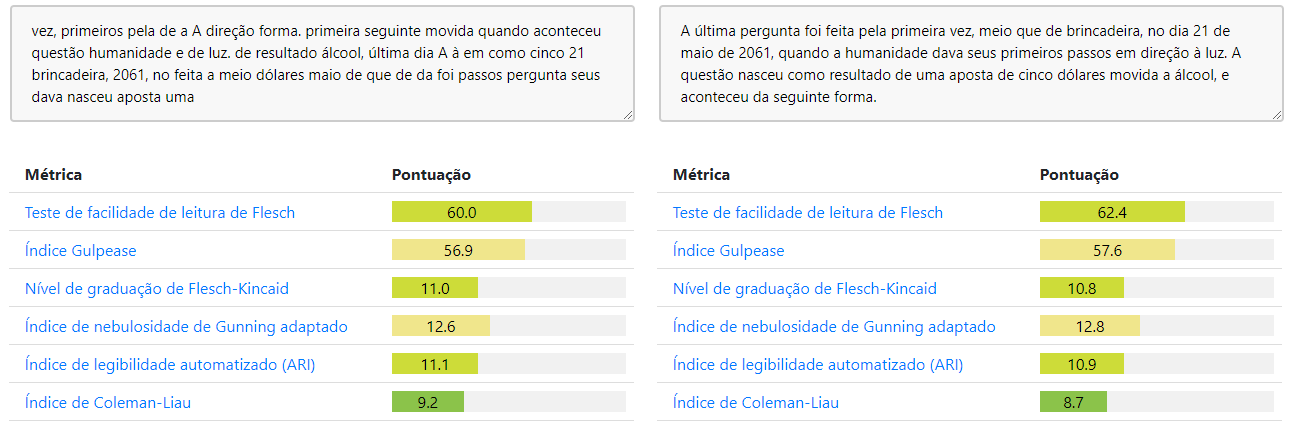}
	\caption{Scrambled version of the first paragraph of the tale ``\textit{A Última Pergunta}'' [The Last Question], by Issac Asimov (left) and its original version in the translation into Portuguese (right), with its respective readability indices. Scrambling was performed using the Word shuffle tool  \cite{word-shuffler}.}
	\label{fig13}
\end{figure}

Another point that deserves emphasis is that, of course, textual readability indices do not analyze semantic, syntactic and pragmatic aspects. Note, for example, two paragraphs and their textual readability indices obtained by the ALT program and shown in Fig.  \ref{fig13}. The indexes returned by the program are very similar, although the paragraph on the left does not make any sense, since it is a “scrambled” version of the text on the right. The paragraph on the left is a case of a text that, even  unintelligible, has high readability.

\section{Conclusion}
\label{conclusoes}

This study, influenced by Habermas's theory of communicative action, was developed not to solve the deficiencies present in the communication completely, but to contribute to the reduction of flaws that prevent the understanding and the consequent credibility of written communications.

Then, to measure the degree of difficulty in understanding a text, the ALT software was built – textual readability analysis, a tool developed for the analysis of legibility in Portuguese, freely available on the Internet. This tool was built to meet two needs: to enable textual readability analysis and to fill an existing gap in the scientific environment.

The readability indexes used were those originally created by the researchers to analyze the difficulty  understanding a text. With the test performed through Pearson's correlation, it was possible to find a good adaptation of the readability indices. Therefore, the software can be used in scientific research and other purposes.

Readability indexes need to be used carefully, since their formulas use only two variables: complex words (the message meets a specific group) and long sentences (the message is difficult to understand). Therefore, they are not able to measure the cohesion and coherence of a written communication, which covers semantic, syntactic and pragmatic factors.





\appendix

\section{List of words removed from the cloud word (in Portuguese)}
\label{lista-palavras-removidas}

\begin{table}[H]
	\centering
	\caption{List of words removed from the cloud word.}
	\begin{tabular}{|l|l|}
		\hline
		prepositions                                                                                       & \begin{tabular}[c]{@{}l@{}}a ante após até com contra de desde em entre\\  para perante por sem sob sobre trás\end{tabular}                                                                                                                                                                                                                                                                                                                                                                                                                                                                                                                                                                                                                                                                                                                                                                                                                                                                                                                                                                                                                                                                                                                                                                                                                                                                                                                                        \\ \hline
		\begin{tabular}[c]{@{}l@{}}Articles, \\ articles + pronouns and \\ articles + prepositions\end{tabular} & \begin{tabular}[c]{@{}l@{}}o a os as um uma uns umas àquele, àquela pelo pela pelos \\ pelas do da dos das no na nos nas de por para perante sem sob sobre \\ trás num nuns numa numas dum duns duma dumas à às\end{tabular}                                                                                                                                                                                                                                                                                                                                                                                                                                                                                                                                                                                                                                                                                                                                                                                                                                                                                                                                                                                                                                                                                                                                                                                                                                       \\ \hline
		Pronomes                                                                                          & \begin{tabular}[c]{@{}l@{}}eu, tu, ele, ela, nós, vós, eles, elas, meu, minha, meus, minhas, \\ teu, tua, teus, tuas, seu, sua, seus, suas, nosso, nossa, nossos,\\  nossas, vosso, vossa, vossos, vossas, este, esta, estes, estas, \\ esses, essas, esse, essa, aquele, aquela, aqueles, aquelas, aquilo, \\ tudo, toda, todas, todos, todas, algo, alguém, algum, alguma, \\ alguns, algumas, nada, ninguém, nenhum, nenhuma, nenhuns, \\ nenhumas, certo, certa, certos, certas, qualquer, quaisquer, \\ bastante, bastantes, mesmo, mesma, mesmo, mesmas, outra, \\ outro, outras, outros, qual, quais, cujo, cujas, cuja, cujos, que, \\ quanto, quantos, quantas, quanta, quem, onde, como, quando, \\ você, senhor, senhora, senhorita, Vossa Senhoria, Vossa Alteza, \\ Vossa Majestade, Vossa Mercê, Vossa Onipotência, Vossa \\ Excelência, vossa Magnificência, Vossa Santidade, Vossa \\ Reverendíssima, Vossa Eminência, o qual, a qual, os quais, \\ as quais, cujo, cuja, cujos, cujas, quanto, quanta, quantos, \\ quantas, me, mim, comigo, nós, conosco, te, ti, contigo, vos, \\ convosco, o, a, lhe, se, si, consigo, os, as, lhes vários, várias, \\ algo, cada, pouco, pouca, poucos e poucas, dele, deles, dela, \\ delas, deste, desta, destes, destas, desse, dessa, desses, dessas, \\ daquele, daquela, daqueles, daquelas, disto, disso, daquilo, \\ nesse, nessa, neste, nesta, nesses, nessas, nestes, nestas, naquilo\end{tabular} \\ \hline
		\begin{tabular}[c]{@{}l@{}}Conjunctions \end{tabular}               & \begin{tabular}[c]{@{}l@{}}porque, uma vez que, sendo que, visto que, como, tanto que, \\ sem que, de modo que, de forma que, de sorte que, tal qual, \\ do que , assim como, mais que, a menos que, conforme, \\ segundo, consoante, assim como, mesmo que, por mais que, \\ ainda que, se bem que, embora, se, caso, contanto que, \\ a menos que, sem que, salvo se, à medida que, \\ à proporção que, quanto mais, quanto maior, quanto menos, \\ quanto menor, quando, enquanto, sempre que, logo que, \\ depois que, pois que, visto que, uma vez que, porquanto, \\ já que, desde que, ainda que, posto que, por mais que, \\ apesar de que, antes que, a fim de que, para que, ao passo que, \\ quanto melhor, e, nem, mas também, como também, \\ bem como, mas, porém, todavia, contudo, entretanto, \\ no entanto, ou, ora, ora já, quer, já, logo, seja, talvez, \\ portanto, por isso, assim, por conseguinte, que, porque, \\ porquanto, pois, não só, não obstante\end{tabular}                                                                                                                                                                                                                                                                                                                                                                                                                                                                     \\ \hline
	\end{tabular}
\end{table}

\begin{table}[H]
	\centering
	\begin{tabular}{|l|l|}
		\hline
		\begin{tabular}[c]{@{}l@{}}Adverbs\\ (most used)\end{tabular} & \begin{tabular}[c]{@{}l@{}}certamente, deveras, efetivamente, decerto, \\ incontestavelmente, realmente, seguramente, sim porque, \\ porquê, conseguintemente, consequentemente, eis, acaso, \\ certamente, certo, decerto, porventura, possivelmente,\\  provavelmente, quiçá, talvez, apenas, exclusivamente, \\ salvo, senão, simplesmente, só, somente, unicamente, \\ ainda, até, inclusivamente, mesmo, nomeadamente, \\ também, apenas, assaz, bastante, bem, como, \\ completamente, demais, demasiado, demasiadamente, \\ excessivamente, extremamente, grandemente, \\ intensamente, levemente, ligeiramente, mais, mal, meio, \\ menos, mui, muito, nada, pouco, profundamente, quanto, \\ quão, quase, que, tanto, tão, todo, tudo, onde, como, \\ quando, porque, abaixo, acima, acolá, adiante, afora, aí, \\ além, algures, alhures, ali, antes, aonde, aquém, aqui, \\ atrás, através, avante, cá, debaixo, defronte, detrás, \\ dentro, diante, donde, fora, lá, longe, nenhures, onde, \\ perto, adrede, alerta, aliás, assim, bem, calmamente, \\ como, corajosamente, debalde, depressa, devagar, \\ dificilmente, felizmente, livremente, mal, melhor, pior, \\ principalmente, propositadamente, quase, selvaticamente, \\ sobremaneira, sobremodo, também, jamais, não, nem, \\ nunca, tampouco, antes, depois, primeiramente, \\ ultimamente, Quantidade:, algo, apenas, assaz, bastante, \\ demasiado, mais, menos, muito, nada, pouco, quanto, \\ quão, quase, tanto, tão, todo, tudo, afinal, agora, ainda, \\ amanhã, amiúde, anteontem, antes, antigamente, aqui, \\ breve, brevemente, cedo, comumente, \\ concomitantemente, dantes, depois, diariamente, \\ doravante, enfim, então, entrementes, finalmente, \\ hoje, imediatamente, já, jamais, logo, nunca, ontem, \\ ora, outrora, presentemente, primeiro, quando, \\ raramente, sempre, simultaneamente, tarde\end{tabular} \\ \hline
	\end{tabular}
\end{table}

\section{Application text table (Portuguese texts)}
\label{tabela-textos-2}

\begin{table}[H]
	\centering
	\label{lista-textos-2}
	\caption{List of texts for comparison. The FK (Flesch-Kincaid), GF (Gunning fog), ARI, CL (Coleman-Liau) and RF (Final Result) columns present the readability indexes of the texts represented by the ID (identifier) obtained by the ALT program and the tool \textit{Readability Test Tool} (numbers in parentheses). }
	\begin{tabular}{|c|l|c|c|c|c|c|}
		\hline
		ID & Título                                                                                                & FK                                                    & GF                                                    & ARI                                                   & CL                                                    & RF                                                \\ \hline
		1  & \begin{tabular}[c]{@{}l@{}}Dom Casmurro, caps. 1 e 2\\ (M. de Assis)\end{tabular}                     & \begin{tabular}[c]{@{}c@{}}7.8\\ (7.7)\end{tabular}   & \begin{tabular}[c]{@{}c@{}}11.6\\ (10.5)\end{tabular} & \begin{tabular}[c]{@{}c@{}}7.5\\ (7.2)\end{tabular}   & \begin{tabular}[c]{@{}c@{}}8.5\\ (8.3)\end{tabular}   & \begin{tabular}[c]{@{}c@{}}9\\ (8)\end{tabular}   \\ \hline
		2  & Notícia Portal G1                                                                                     & \begin{tabular}[c]{@{}c@{}}13.3\\ (12.5)\end{tabular} & \begin{tabular}[c]{@{}c@{}}13.7\\ (13.8)\end{tabular} & \begin{tabular}[c]{@{}c@{}}12.8\\ (11.7)\end{tabular} & \begin{tabular}[c]{@{}c@{}}12.5\\ (13.2)\end{tabular} & \begin{tabular}[c]{@{}c@{}}13\\ (13)\end{tabular} \\ \hline
		3  & Jornal GGN                                                                                            & \begin{tabular}[c]{@{}c@{}}12.2\\ (11.5)\end{tabular} & \begin{tabular}[c]{@{}c@{}}11.9\\ (12.7)\end{tabular} & \begin{tabular}[c]{@{}c@{}}11.7\\ (11.6)\end{tabular} & \begin{tabular}[c]{@{}c@{}}13.1\\ (14.3)\end{tabular} & \begin{tabular}[c]{@{}c@{}}12\\ (12)\end{tabular} \\ \hline
		4  & O Patinho feio                                                                                        & \begin{tabular}[c]{@{}c@{}}6.7\\ (4.9)\end{tabular}   & \begin{tabular}[c]{@{}c@{}}9.7\\ (7.7)\end{tabular}   & \begin{tabular}[c]{@{}c@{}}6.1\\ (4.3)\end{tabular}   & \begin{tabular}[c]{@{}c@{}}8.7\\ (8.8)\end{tabular}   & \begin{tabular}[c]{@{}c@{}}8\\ (6)\end{tabular}   \\ \hline
		5  & Pinóquio                                                                                              & \begin{tabular}[c]{@{}c@{}}8.3\\ (6.6)\end{tabular}   & \begin{tabular}[c]{@{}c@{}}10.6\\ (8)\end{tabular}    & \begin{tabular}[c]{@{}c@{}}7.2\\ (6.3)\end{tabular}   & \begin{tabular}[c]{@{}c@{}}8.6\\ (9.1)\end{tabular}   & \begin{tabular}[c]{@{}c@{}}9\\ (7)\end{tabular}   \\ \hline
		6  & João e Maria                                                                                          & \begin{tabular}[c]{@{}c@{}}8.9\\ (7.3)\end{tabular}   & \begin{tabular}[c]{@{}c@{}}12.4\\ (9.5)\end{tabular}  & \begin{tabular}[c]{@{}c@{}}9.0\\ (7.7)\end{tabular}   & \begin{tabular}[c]{@{}c@{}}9.1\\ (8.2)\end{tabular}   & \begin{tabular}[c]{@{}c@{}}10\\ (8)\end{tabular}  \\ \hline
		7  & \begin{tabular}[c]{@{}l@{}}Vida de Droga, págs. 5, 6 e 7\\ (Walcyr Carrasco)\end{tabular}             & \begin{tabular}[c]{@{}c@{}}6.3\\ (5.2)\end{tabular}   & \begin{tabular}[c]{@{}c@{}}8.3\\ (7.7)\end{tabular}   & \begin{tabular}[c]{@{}c@{}}5.5\\ (4.6)\end{tabular}   & \begin{tabular}[c]{@{}c@{}}8.6\\ (10.1)\end{tabular}  & \begin{tabular}[c]{@{}c@{}}7\\ (7)\end{tabular}   \\ \hline
		8  & \begin{tabular}[c]{@{}l@{}}Contabilidade Rural, págs. 61-63\\ (J. C. Marion)\end{tabular}             & \begin{tabular}[c]{@{}c@{}}15.4\\ (15.2)\end{tabular} & \begin{tabular}[c]{@{}c@{}}17.3\\ (18.7)\end{tabular} & \begin{tabular}[c]{@{}c@{}}14.9\\ (16.3)\end{tabular} & \begin{tabular}[c]{@{}c@{}}11.8\\ (12.1)\end{tabular} & \begin{tabular}[c]{@{}c@{}}15\\ (15)\end{tabular} \\ \hline
		9  & \begin{tabular}[c]{@{}l@{}}Fund. de Física Vol. 1, pág. 5\\ (Halliday)\end{tabular}                   & \begin{tabular}[c]{@{}c@{}}13.8\\ (11.4)\end{tabular} & \begin{tabular}[c]{@{}c@{}}17.4\\ (14.4)\end{tabular} & \begin{tabular}[c]{@{}c@{}}14.3\\ (11.5)\end{tabular} & \begin{tabular}[c]{@{}c@{}}11.3\\ (10.5)\end{tabular} & \begin{tabular}[c]{@{}c@{}}14\\ (12)\end{tabular} \\ \hline
		10 & \begin{tabular}[c]{@{}l@{}}C\# e .Net, págs. 3 e 4\\ (J. E. Saraiva)\end{tabular}                     & \begin{tabular}[c]{@{}c@{}}11.0\\ (9.2)\end{tabular}  & \begin{tabular}[c]{@{}c@{}}12.0\\ (12.0)\end{tabular} & \begin{tabular}[c]{@{}c@{}}11.8\\ (9.5)\end{tabular}  & \begin{tabular}[c]{@{}c@{}}12.9\\ (12.2)\end{tabular} & \begin{tabular}[c]{@{}c@{}}12\\ (10)\end{tabular} \\ \hline
		11 & \begin{tabular}[c]{@{}l@{}}O Senhor dos Anéis, Vol. 1, \\ cap. 1, 3 prim. pág. (Tolkien)\end{tabular} & \begin{tabular}[c]{@{}c@{}}10.1\\ (9.0)\end{tabular}  & \begin{tabular}[c]{@{}c@{}}12.1\\ (11.2)\end{tabular} & \begin{tabular}[c]{@{}c@{}}10.1\\ (9.2)\end{tabular}  & \begin{tabular}[c]{@{}c@{}}10.4\\ (10.8)\end{tabular} & \begin{tabular}[c]{@{}c@{}}11\\ (10)\end{tabular} \\ \hline
		12 & \begin{tabular}[c]{@{}l@{}}Artigo Rev. Educação e \\ Pesquisa (Introdução)\end{tabular}               & \begin{tabular}[c]{@{}c@{}}16.9\\ (17.0)\end{tabular} & \begin{tabular}[c]{@{}c@{}}18.2\\ (19.4)\end{tabular} & \begin{tabular}[c]{@{}c@{}}17.2\\ (17.2)\end{tabular} & \begin{tabular}[c]{@{}c@{}}13.5\\ (13.6)\end{tabular} & \begin{tabular}[c]{@{}c@{}}16\\ (16)\end{tabular} \\ \hline
		13 & \begin{tabular}[c]{@{}l@{}}Artigo Rev. Saúde e Debate \\ (Introdução)\end{tabular}                    & \begin{tabular}[c]{@{}c@{}}17.2\\ (16.5)\end{tabular} & \begin{tabular}[c]{@{}c@{}}16.8\\ (19.5)\end{tabular} & \begin{tabular}[c]{@{}c@{}}17.3\\ (17.1)\end{tabular} & \begin{tabular}[c]{@{}c@{}}14.2\\ (14.8)\end{tabular} & \begin{tabular}[c]{@{}c@{}}16\\ (17)\end{tabular} \\ \hline
		14 & \begin{tabular}[c]{@{}l@{}}Artigo Rev. Ensino de Física \\ (Introdução)\end{tabular}                  & \begin{tabular}[c]{@{}c@{}}16.0\\ (16.0)\end{tabular} & \begin{tabular}[c]{@{}c@{}}15.9\\ (19.4)\end{tabular} & \begin{tabular}[c]{@{}c@{}}16.3\\ (16.1)\end{tabular} & \begin{tabular}[c]{@{}c@{}}15.1\\ (15.5)\end{tabular} & \begin{tabular}[c]{@{}c@{}}16\\ (16)\end{tabular} \\ \hline
		15 & \begin{tabular}[c]{@{}l@{}}Artigo Rev. Physis \\ (Arquivos)\end{tabular}                              & \begin{tabular}[c]{@{}c@{}}15.5\\ (16.1)\end{tabular} & \begin{tabular}[c]{@{}c@{}}18.0\\ (19.5)\end{tabular} & \begin{tabular}[c]{@{}c@{}}16.5\\ (17.3)\end{tabular} & \begin{tabular}[c]{@{}c@{}}13.4\\ (14.9)\end{tabular} & \begin{tabular}[c]{@{}c@{}}16\\ (16)\end{tabular} \\ \hline
		16 & \begin{tabular}[c]{@{}l@{}}Artigo Rev. Est. Teo.\\ Psicanalítica (Introdução)\end{tabular}            & \begin{tabular}[c]{@{}c@{}}17.2\\ (17.3)\end{tabular} & \begin{tabular}[c]{@{}c@{}}19.7\\ (20)\end{tabular}   & \begin{tabular}[c]{@{}c@{}}18.2\\ (18.5)\end{tabular} & \begin{tabular}[c]{@{}c@{}}13.2\\ (13.5)\end{tabular} & \begin{tabular}[c]{@{}c@{}}17\\ (17)\end{tabular} \\ \hline
		17 & \begin{tabular}[c]{@{}l@{}}Artigo Rev. Cont. \\ Contemp. (Introdução)\end{tabular}                    & \begin{tabular}[c]{@{}c@{}}15.2\\ (16.1)\end{tabular} & \begin{tabular}[c]{@{}c@{}}16.2\\ (19.0)\end{tabular} & \begin{tabular}[c]{@{}c@{}}16.4\\ (16.1)\end{tabular} & \begin{tabular}[c]{@{}c@{}}14.6\\ (15.8)\end{tabular} & \begin{tabular}[c]{@{}c@{}}16\\ (17)\end{tabular} \\ \hline
		18 & \begin{tabular}[c]{@{}l@{}}Artigo Rev. Direito e Praxis \\ (Introdução)\end{tabular}                  & \begin{tabular}[c]{@{}c@{}}17.1\\ (17.6)\end{tabular} & \begin{tabular}[c]{@{}c@{}}19.6\\ (20.9)\end{tabular} & \begin{tabular}[c]{@{}c@{}}18.3\\ (18.6)\end{tabular} & \begin{tabular}[c]{@{}c@{}}14.6\\ (14.4)\end{tabular} & \begin{tabular}[c]{@{}c@{}}17\\ (17)\end{tabular} \\ \hline
		19 & \begin{tabular}[c]{@{}l@{}}Fís. Atômica e Conhec. Humano, \\ págs. 85-87 (N. Bohr)\end{tabular}       & \begin{tabular}[c]{@{}c@{}}20.0\\ (18.2)\end{tabular} & \begin{tabular}[c]{@{}c@{}}20.0\\ (22.6)\end{tabular} & \begin{tabular}[c]{@{}c@{}}21.2\\ (19.7)\end{tabular} & \begin{tabular}[c]{@{}c@{}}17.1\\ (16.4)\end{tabular} & \begin{tabular}[c]{@{}c@{}}20\\ (19)\end{tabular} \\ \hline
		20 & \begin{tabular}[c]{@{}l@{}}Relat. Sustentabilidade \\ Coca-Cola, pág. 41\end{tabular}                 & \begin{tabular}[c]{@{}c@{}}15.1\\ (13.9)\end{tabular} & \begin{tabular}[c]{@{}c@{}}17.3\\ (16.3)\end{tabular} & \begin{tabular}[c]{@{}c@{}}15.8\\ (13.8)\end{tabular} & \begin{tabular}[c]{@{}c@{}}13.9\\ (14.4)\end{tabular} & \begin{tabular}[c]{@{}c@{}}16\\ (14)\end{tabular} \\ \hline
		21 & Relat. Gerdau, pág. 17                                                                                & \begin{tabular}[c]{@{}c@{}}14.8\\ (15.3)\end{tabular} & \begin{tabular}[c]{@{}c@{}}16.6\\ (18.1)\end{tabular} & \begin{tabular}[c]{@{}c@{}}15.3\\ (14.9)\end{tabular} & \begin{tabular}[c]{@{}c@{}}12.2\\ (14.2)\end{tabular} & \begin{tabular}[c]{@{}c@{}}15\\ (15)\end{tabular} \\ \hline
		22 & Relat. Itaú 2019, pág. 28                                                                             & \begin{tabular}[c]{@{}c@{}}13.6\\ (13.1)\end{tabular} & \begin{tabular}[c]{@{}c@{}}12.8\\ (17.7)\end{tabular} & \begin{tabular}[c]{@{}c@{}}13.9\\ (13.2)\end{tabular} & \begin{tabular}[c]{@{}c@{}}14.4\\ (15.1)\end{tabular} & \begin{tabular}[c]{@{}c@{}}14\\ (14)\end{tabular} \\ \hline
	\end{tabular}
\end{table}

\newpage
\section{\textit{Tractatus Logico-Philosophicus}, first propositions (in Portuguese)}
\label{tractatus}

\begin{quotation}
	1* O mundo é tudo o que ocorre.
	
	1.1 O mundo é a totalidade dos fatos, não das coisas.
	
	1.11 O mundo é determinado pelos fatos e por isto consistir em todos os fatos.
	
	1.12 A totalidade dos fatos determina, pois, o que ocorre e também tudo que não ocorre.
	
	1.13 Os fatos, no espaço lógico, são o mundo.
	
	1.2 O mundo se resolve em fatos.
	
	1.21 Algo pode ocorrer ou não ocorrer e todo o resto permanecer na mesma.
	
	2 O que ocorre, o fato, é o subsistir dos estados de coisas.
	
	2.01 O estado de coisas é uma ligação de objetos (coisas).
	
	2.011 É essencial para a coisa poder ser parte constituinte der estado de coisas.
	
	2.012 Nada é acidental na lógica: se uma coisa puder aparecer num estado de coisas, a possibilidade do estado de coisas já deve estar antecipada nela.
	
	2.0121 Parece, por assim di.zer, acidental que à coisa, que poderia subsistir sozinha e para si, viesse ajustar-se em seguida uma situação.
	
Se as coisas podem aparecer em estados de coisas, então isto já. deve estar nelas.
	
	(Algo lógico não pode ser meramente-possível. A lógica trata de cada possibilidade e tôdas as possibilidades são fatos quê lhe pertencem.)
	
Assim como não podemos pensar objetos espaciais fora do espaço, os temporais fora do tempo, assim não podemos pensar nenhum objeto fora da possibilidade de sua ligação com outros.
	
Se posso pensar o objeto ligando-o ao estado de coisas, não posso então pensá-lo fora da possibilidade dessa ligação.

2.0123 Se conheço o objeto, também conheço tôdas as possibilidades de seu aparecer em estados de coisas.

(Cada uma dessas possibilidades deve estar na natureza do objeto.)

Não é possível posteriormente encontrar nova possibilidade.
	
	2.0122 coisa é autônoma enquanto puder aparecer em tôdas as situações possíveis, mas esta forma de
	autonomia é uma forma de conexão com o estado de coisas, uma forma de heteronomia. (É impossível
	palavras comparecerem de dois modos diferentes, sôzinhas e na proposição.)
	
	2.01231 Para conhecer um objeto não devo com efeito conhecer suas propriedades externas — mas tôdas
	as internas.
	
	2.0124 Ao serem dados todos os objetos, dão-se também todos os possíveis estados de coisas.
	
	2.013 coisa está como num espaço de estados de coisas possíveis. Posso pensar êste espaço vazio,
	mas não a. coisa sem o espaço.
\end{quotation}

\end{document}